\documentclass[conference]{IEEEtran}
\IEEEoverridecommandlockouts
\usepackage{cite}
\usepackage{amsmath,amssymb,amsfonts}
\usepackage[colorlinks=true, linkcolor=black, urlcolor=black]{hyperref}
\usepackage{algorithmic}
\usepackage{graphicx}
\usepackage{textcomp}
\usepackage{xcolor}
\usepackage{amsmath,amsfonts}
\usepackage{algorithmic}
\usepackage{algorithm}
\usepackage{array}
\usepackage{caption}
\usepackage[caption=false,font=normalsize,labelfont=sf,textfont=sf]{subfig}
\usepackage{textcomp}
\usepackage{authblk}
\usepackage{graphicx}  
\usepackage{lipsum}    
\usepackage{float}
\usepackage{booktabs}
\usepackage{stfloats}
\usepackage{url}
\usepackage{verbatim}
\usepackage{graphicx}
\usepackage{cite}
\usepackage{balance}
\usepackage{makecell}
\usepackage{enumitem}
\usepackage{multirow}
\usepackage{pifont}
\usepackage{colortbl}
\usepackage{subcaption}
\def\BibTeX{{\rm B\kern-.05em{\sc i\kern-.025em b}\kern-.08em
    T\kern-.1667em\lower.7ex\hbox{E}\kern-.125emX}}
\begin{document}

\title{MoDE: Mixture of Diffusion Experts for Any Occluded Face Recognition}
%\\

% \author{
%     Qiannan Fan, Zhuoyang Li, Jitong Li, Chenyang Cao \\
%     College of Intelligence and Computing, Tianjin University, Tianjin, China \\
%     qiannanfan@outlook.com, 2125867922@qq.com, {jitong\_li@yeah.net}, 1035363213@qq.com }
\author{
Qiannan Fan$^{1,2}$, Zhuoyang Li$^{2}$, Jitong Li$^{2\dag}$, Chenyang Cao$^{2*}$ \\
$^{1}$State Grid Tianjin Economic Research Institute, Tianjin, China \\
$^{2}$College of Intelligence and Computing, Tianjin University, Tianjin, China \\
qiannanfan@outlook.com, 2125867922@qq.com, jitong\_li@yeah.net, 1035363213@qq.com
}

\maketitle
\begingroup
\renewcommand\thefootnote{}\footnotetext{\textsuperscript{\dag}Work done during internship at Tianjin University.}
\footnotetext{\textsuperscript{*}Corresponding author.}
\endgroup
% \vspace{-3ex}
% \begin{figure}[H]
% \hsize=\textwidth 
% \centering
% \includegraphics[width=\textwidth]{figure/Figure0.jpg}
% \caption{To demonstrate the adaptability of our method for multi-occluded face recognition, we visualize some experimental results. The baseline model is trained on MS1M with the ResNet-50 architecture and ArcFace loss. The left part visualizes face images on the LFW dataset with six different face occlusions, each image is repainted below. The right part visualizes the accuracy comparison between the baseline and our method (MoDE). Note that the legends on the left indicate the normal face and different occlusion categories (e.g., Ground Truth represents the normal face, and Mask represents mask occlusion).}
% \label{fig:teaser}
% \end{figure}
% }

\begin{abstract}
With the continuous impact of epidemics, people have become accustomed to wearing masks. However, most current occluded face recognition (OFR) algorithms lack prior knowledge of occlusions, resulting in poor performance when dealing with occluded faces of varying types and severity in reality. Recognizing occluded faces is still a significant challenge, which greatly affects the convenience of people's daily lives. In this paper, we propose an identity-gated mixture of diffusion experts (MoDE) for OFR. Each diffusion-based generative expert estimates one possible complete image for occluded faces. Considering the random sampling process of the diffusion model, which introduces inevitable differences and variations between the inpainted faces and the real ones. To ensemble effective information from multi-reconstructed faces, we introduce an identity-gating network to evaluate the contribution of each reconstructed face to the identity and adaptively integrate the predictions in the decision space. Moreover, our MoDE is a plug-and-play module for most existing face recognition models. Extensive experiments on three public face datasets and two datasets in the wild validate our advanced performance for various occlusions in comparison with the competing methods.
\end{abstract}

\begin{IEEEkeywords}
occluded face recognition, generative model, diffusion expert
\end{IEEEkeywords}

\section{Introduction}
Face recognition is ubiquitous and has been widely applied in various applications~\cite{kortli2020face}, such as security, payment, and social media. Recent face recognition algorithms~\cite{sun2014deep,zhao20183d}, such as FaceNet~\cite{schroff2015facenet,cao2018vggface2}, CosFace~\cite{wang2018cosface}, and ArcFace~\cite{deng2019arcface}, have shown remarkable performance in unobstructed face recognition. However, many practical scenarios may cause the face to be obscured. In particular, the frequent flu epidemics have led to more people getting used to wearing masks to protect themselves and others, bringing new challenges to face recognition, which significantly impacts the convenience of daily security. Moreover, the uncertainty of occluded face regions and the diversity of occlusion types (\textit{e.g.}, accessories, glasses, hats, and masks) require more robust and generalized face recognition techniques~\cite{cao2018asymmetric}. The occluded face parts usually result in discriminative information loss and feature changes, leading to a significant decline in the recognition accuracy of most traditional face recognition models.
\begin{figure}[t]
% \hsize=0.5\textwidth 
\centering
\includegraphics[width=0.5\textwidth]{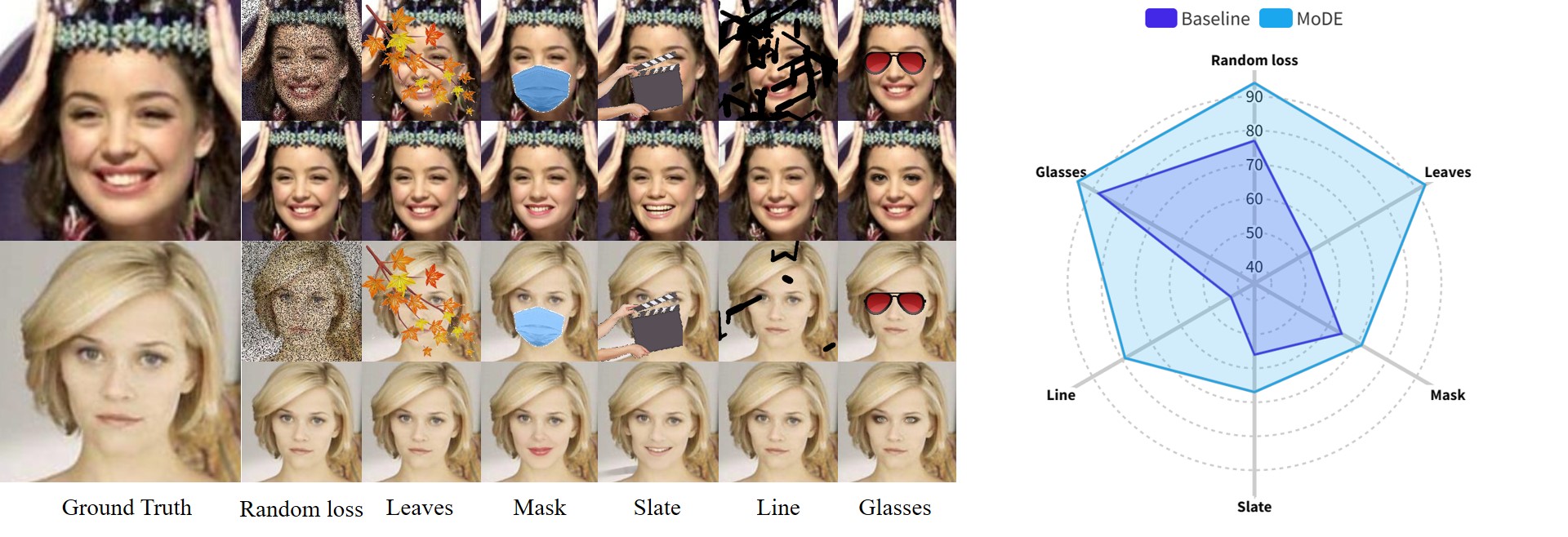}
\caption{
To demonstrate the adaptability of our method for multi-occluded face recognition, 
we visualize some 
% experimental results. The baseline model is trained on MS1M with the ResNet-50 architecture and ArcFace loss. The left part visualizes 
face images from the LFW dataset with six different face occlusions, each image is repainted below. The right part shows the accuracy comparison between the baseline and our method. 
% Note that the legends on the left indicate the normal face and different occlusion categories (e.g., Ground Truth represents the normal face, and Mask represents mask occlusion).
}
\label{fig:teaser}
\end{figure}
% \notsure{Studies such as Li et al. ~\cite{li2018occlusion} and Azeem et al. ~\cite{azeem2014survey} have explored the impact of occlusion and surveyed the effect of masks on face recognition accuracy.}

In recent years, some studies have emerged to improve occluded face recognition accuracy. These methods can be mainly grouped into three categories~\cite{zeng2021survey}. i) Occlusion robust feature extraction (ORFE) aims to extract features that are less affected by occlusions, such as Local Binary Patterns (LBP)~\cite{ahonen2004face,ahonen2006face} and Elastic Bunch Graph Matching (EBGM)~\cite{zhao2015modular}. However, most of these methods rely on handcrafted features and require prior knowledge, which limits their effectiveness. ii) Occlusion-aware face recognition (OAFR) discards occluded parts and only uses visible face parts that qualify for face recognition. This approach requires detecting the occluded parts, which can be challenging in the case of irregular occlusions. iii) Occlusion recovery-based face recognition (ORecFR) uses a reconstruction model to repair occluded faces for the demands of conventional face recognition systems. However, these models may not generalize well to occlusions that are not present in the training set, which could result in unsatisfied results~\cite{cao2018data}. Furthermore, while the reconstructed images may present visually convincing, they might be tangled in poor authenticity and identity ambiguity.

% lzydelete
% \vspace{-3ex}
% \begin{figure}[t]
% \centering
% \includegraphics[width=0.47\textwidth]{figure/Figure0.jpg}
% \caption{To demonstrate the adaptability of our method for multi-occluded face recognition, we visualize some experimental results. The baseline model is trained on MS1M with the ResNet-50 architecture and ArcFace loss. The left part visualizes face images on the LFW dataset with six different face occlusions, each image is repainted below. The right part visualizes the accuracy comparison between the baseline and our method (MoDE). Note that the legends on the left indicate the normal face and different occlusion categories (e.g., Ground Truth represents the normal face, and Mask represents mask occlusion).}
% \label{fig:teaser}
% \vspace{-24px}
% \end{figure}

In this paper, we propose an identity-gated mixture of diffusion experts (MoDE) framework for occluded face recognition. 
% Our model combines the Mixture of Experts (MoE)~\cite{shazeer2017outrageously} and Denoising Diffusion Probabilistic models (DDPM)~\cite{ho2020denoising}.  
By integrating multiple diffusion experts, we obtain identity-preserving reconstructed face images while improving recognition accuracy. Specifically, the MoDE utilizes diffusion experts to generate reconstructed images for each occluded face, regardless of the type of occlusions. 
The diffusion model endows diverse potential visual appearances for the repaint results that provide both effective and redundant information for identification. To fully exploit the identity information from various reconstructed images, we further design an identity gating (ID-Gate) network that estimates the authenticity of multiple input images and outputs the corresponding contribution to the identification of the similarity matrix. To this end, MoDE exhibits excellent performance for multiple occluded face recognition, Fig. \ref{fig:teaser} illustrates it. Furthermore, the adaptively selected results of MoDE serve as valuable references for evaluating the identity consistency of the reconstructed images.
We have performed experiments on various datasets to demonstrate the superiority against competing methods.
% .Fig. \ref{fig:teaser} further illustrates our adaptability and flexibility for multi-occluded face recognition scenarios. 
Our main contributions can be summarized in three aspects:
\begin{itemize}
% [itemsep= 1pt, topsep = 1pt]
\vspace{-2pt}
\item [1)] We propose a dynamic occluded face recognition framework, which integrates multiple generative experts in the decision space through ID-Gate and subtly leverages comprehensive identity information produced by them. 
% This framework provides a new avenue for practical occluded face recognition. 
\item [2)] We introduce the diffusion model to reconstruct occluded faces with high image quality and diversity. Our approach is capable of identifying various types of occlusions, even in extreme conditions. 
To our knowledge, our work is the first to introduce the diffusion model to occluded face recognition. 
\item [3)] Our MoDE is a flexible \textit{plug-and-play} module that can be seamlessly combined with most existing face recognition models and achieve stable improvements. Extensive experiments validate our superiority.
% We also validate our superiority in self-collected real occluded face datasets in comparing with the competing methods. Our code will be publicly available and designed to be both user-friendly and easily trainable.
\end{itemize}

\section{Related works}
% \subsection{General Face Recognition}
% Deep face recognition is among the most extensively researched and prosperous domains in pattern recognition. 
% 1
% Recently, deep face recognition has made substantial progress owing to recent advances in deep learning technology, coupled with the availability of large-scale training data sets. 
\noindent\textbf{General Face Recognition} have achieved impressive results and have focused primarily on designing novel network architectures~\cite{zheng2018ring, wen2016discriminative} and loss functions~\cite{zhang2017range,liu2016large}. The early models relied on multiple deep convolutional neural networks to extract facial features, which were later fused together~\cite{taigman2014deepface,cao2020multi}. However, the use of a single network to extract features has now dominated mainly~\cite{schroff2015facenet}. Meanwhile, the softmax loss for multiclass classification is widely used and significantly boosted by angle-based techniques~\cite{liu2017sphereface,wang2018cosface,deng2019arcface}.
% \subsection{Occluded Face Recognition}
% The limitations of face recognition in the presence of occlusions have been a persistent challenge. However, recognizing faces with occlusions is essential for maximizing the potential of face recognition in real-world applications. According to Zeng et al. ~\cite{zeng2021survey}, there are three main types of face recognition algorithms under occlusion: (1) occlusion robust feature extraction (ORFE) methods, (2) occlusion-aware face recognition methods, and (3) face recognition methods based on occlusion recovery.

\begin{figure*}[t]
\centering
\includegraphics[width=1\textwidth]{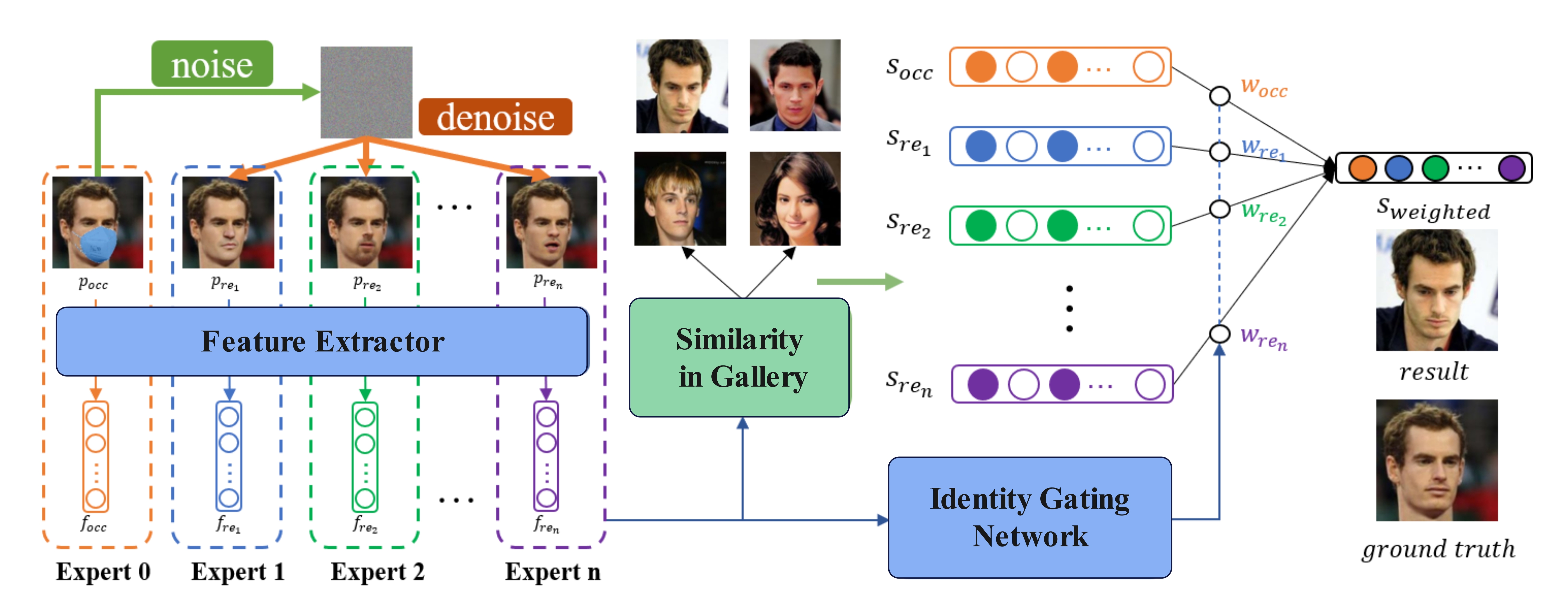}
\vspace{-2pt}
\caption{The framework of our MoDE.
Firstly, MoDE reconstructs the occluded image and produces n repainted images. Then, it extracts the features from the repainted images and the original occluded image, which are then inputted into ID-Gate to output the weight vector. Finally, we calculate the similarity matrix of each image and multiply it with the weight vector to generate a weighted similarity matrix, thereby obtaining the recognition result. 
}
\label{fig1}
\end{figure*}

\noindent\textbf{Occluded Face Recognition} has become a persistent challenge in computer vision~\cite{zeng2021survey}. Existing methods can be broadly categorized into robust occlusion  feature extraction (ORFE), occlusion-aware face recognition (OAFR), and occlusion-recovery-based face recognition (ORecFR). ORFE aims to extract features from an image or video that are robust to occlusions~\cite{wiskott1997face,ahonen2004face,ahonen2006face}. However, the accuracy of these methods is limited by the
limited discriminability of feature representation~\cite {qiu2021end2end}.
OAFR uses visible facial parts alone for recognition to alleviate the occlusion effect~\cite{weng2016robust,he2018dynamic,song2019occlusion}. 
They mainly exploit limited information from partial faces.
ORecFR~\cite{criminisi2004region} recovers the occlusion with non-blind or blind inpainting. 
% For non-blind inpainting techniques, for example, Criminisi et al. ~\cite{criminisi2004region} used sample-based texture synthesis to fill in the damaged regions of an image. Blind inpainting, on the other hand, relies on generative models such as variational autoencoders (VAE) ~\cite{kingma2013auto} or generative adversarial networks (GAN) ~\cite{goodfellow2020generative}. However, when identifying faces
With novel occlusions, these models often illustrate inadequate reconstruction and recognition capabilities.
% lzydelete
% Various GAN-based methods, such as Information Maximizing Generative Adversarial Networks (InfoGAN) ~\cite{chen2016infogan}, Occlusion-aware GAN ~\cite{chen2017occlusion}, AttGAN ~\cite{he2019attgan}, and ID-GAN ~\cite{ge2020occluded}, have been proposed to improve the recognition accuracy of occluded faces. 
% However, when identifying faces with novel occlusions, these models often illustrate inadequate reconstruction and recognition capabilities, which could be attributed to the limited diversity generation capability.

\noindent\textbf{Image Repainting} aims to fill in the gaps and missing components of an image. 
% Traditional approaches to image restoration typically rely on surrounding pixels to estimate missing ones, but are often ill-suited for generating semantic content due to their limited global information~\cite{ge2020occluded}.
% Information Maximizing Generative Adversarial Networks 
% (InfoGAN) 
% ~\cite{chen2016infogan},
% Occlusion-aware GAN ~\cite{chen2017occlusion}
Generative Adversarial Nets (GANs)~\cite{goodfellow2020generative}, such as AttGAN~\cite{he2019attgan}, ID-GAN~\cite{ge2020occluded} have been widely used in various image generation applications~\cite{cao2022face}. 
However, many GAN-based image reconstruction methods~\cite{cao2023autoencoder} often produce deterministic transformations, leading to poor diversity in generated images~\cite{cao2023autoencoder2}, and the restored results may appear seriously ambiguous or blurry~\cite{dhariwal2021diffusion}.
Recently, Diffusion models have gained popularity for their powerful generative capacity to capture complex data distributions and model the underlying dynamics of image generation processes~\cite{sohl2015deep,lugmayr2022repaint,bi2022learning}. Ho et al.~\cite{ho2020denoising} introduced a new SDE-based generative model, Denoising Diffusion Probabilistic models (DDPMs), which employed an incremental generative process to efficiently generate data and capture complex data distributions. 
% Lugmayr et al.~\cite{lugmayr2022repaint} modeled the image data and noise in the damaged region using DDPMs, then solved an optimization problem to reconstruct missing pixels in the manner consistent with the DDPM.
In this paper, we employ the diffusion model to repaint occluded faces, empowering a significant diversity of potential occluded facial components and assembling them for face recognition.
%In this paper, we use the diffusion model for repainting occluded images, as it offers advantages over GANs such as increased diversity and more stable training.

\section{Methodology}
% In this paper, we introduce the MoDE model (as shown in Fig. \ref{fig1}) for face recognition, which repaints the occluded face to generate multiple images as experts and utilizes ID-Gate to weight different experts in the decision space. 
% Firstly, we employ a diffusion model to repaint $n$ facial images from occluded faces, referring to different experts (n+1 in total). Subsequently, their respective features are extracted via a deep feature extraction network. These expert features are utilized to calculate the similarity matrix with registered faces in the gallery. Meanwhile, they are fed to the identity gating network to generate the weight vector, which dynamically fuses experts based on identity authenticity. To this end, the recognition result is obtained by weighting the similarity matrix in the decision space.

% In the subsequent sections, we first detail the diffusion-based face repainting framework. Then, we describe the Identity Gating Network (ID-Gate) and provide a detailed explanation of Mixture of Diffusion Experts (MoDE). We also present the training objectives of our MoDE. Finally, we briefly introduce the network architectures.

\subsection{Face Repainting}

We employ a Denoising Diffusion Probabilistic Model ~\cite{ho2020denoising} in our face repainting framework, as shown in Fig.~\ref{fig1}. It iteratively denoises a random noise sample $x_{T}$ until a high-fidelity output image $x_{0}^{d}$ is obtained. 
% This model predicts the distribution of noise by learning the statistical properties of the images in a specified training set.
During the diffusion process, the initial image $x_{0}^{d}$ diffuses to white Gaussian noise $x_{T}$ over a duration of $T$ time steps.$q\left(x_{t} \mid x_{t-1}\right)$  represents the probability distribution of the current image given the previous image at step 
$t$:
% \begin{equation}  
% x_{t}=\sqrt{1-\beta_{t}} x_{t-1}+\sqrt{\beta_{t}} z.
% \end{equation}
\begin{equation}  
q\left(x_{t} \mid x_{t-1}\right)=\mathcal{N}\left(x_{t} ; \sqrt{1-\beta_{t}} x_{t-1}, \beta_{t} \mathbf{I}\right).
\label{fo1}
\end{equation}
At each step, Gaussian noise with variance $\beta_{t}$ is added to the previous image $x_{t-1}$, preserving $\sqrt{1-\beta_{t}}$ of it, where $\mathbf{I}$ is the identity matrix. 
% \sim \mathcal{N}(0,1)$ is white Gaussian noise.
% Therefore, each step is obeyed as follows:
% \begin{equation}  
% q\left(x_{t} \mid x_{t-1}\right)=\mathcal{N}\left(x_{t} ; \sqrt{1-\beta_{t}} x_{t-1}, \beta_{t} \mathbf{I}\right).
% \end{equation}
By defining $\bar{\alpha}_{t}=\prod_{s=1}^{t}\left(1-\beta_{s}\right)$, we can thus rewrite \ref{fo1}, as a single step:
\begin{equation}  
q\left(x_{t} \mid x_{0}^{d}\right)=\mathcal{N}\left(x_{t} ; \sqrt{\bar{\alpha}_{t}} x_{0}^{d},\left(1-\bar{\alpha}_{t}\right) \mathbf{I}\right).
\end{equation}

For the denoising process, the model predicts the noise $\epsilon_{\theta}$ that is added to $x_{t}$ during the diffusion process, and then subtracts this noise from $x_{t-1}$ to obtain a cleaner image as,
\begin{equation}
p_{\theta}\left(x_{t-1} \mid x_{t}\right)=\mathcal{N}\left(x_{t-1} ; \mu_{\theta}\left(x_{t}, t\right), \Sigma_{\theta}\left(x_{t}, t\right)\right),
\end{equation}
where $\mu_{\theta}\left(x_{t}, t\right)$ and $\Sigma_{\theta}\left(x_{t}, t\right)$ are the parameters of the Gaussian distribution.

\begin{figure}[t]
\centering
\includegraphics[width=0.5\textwidth]{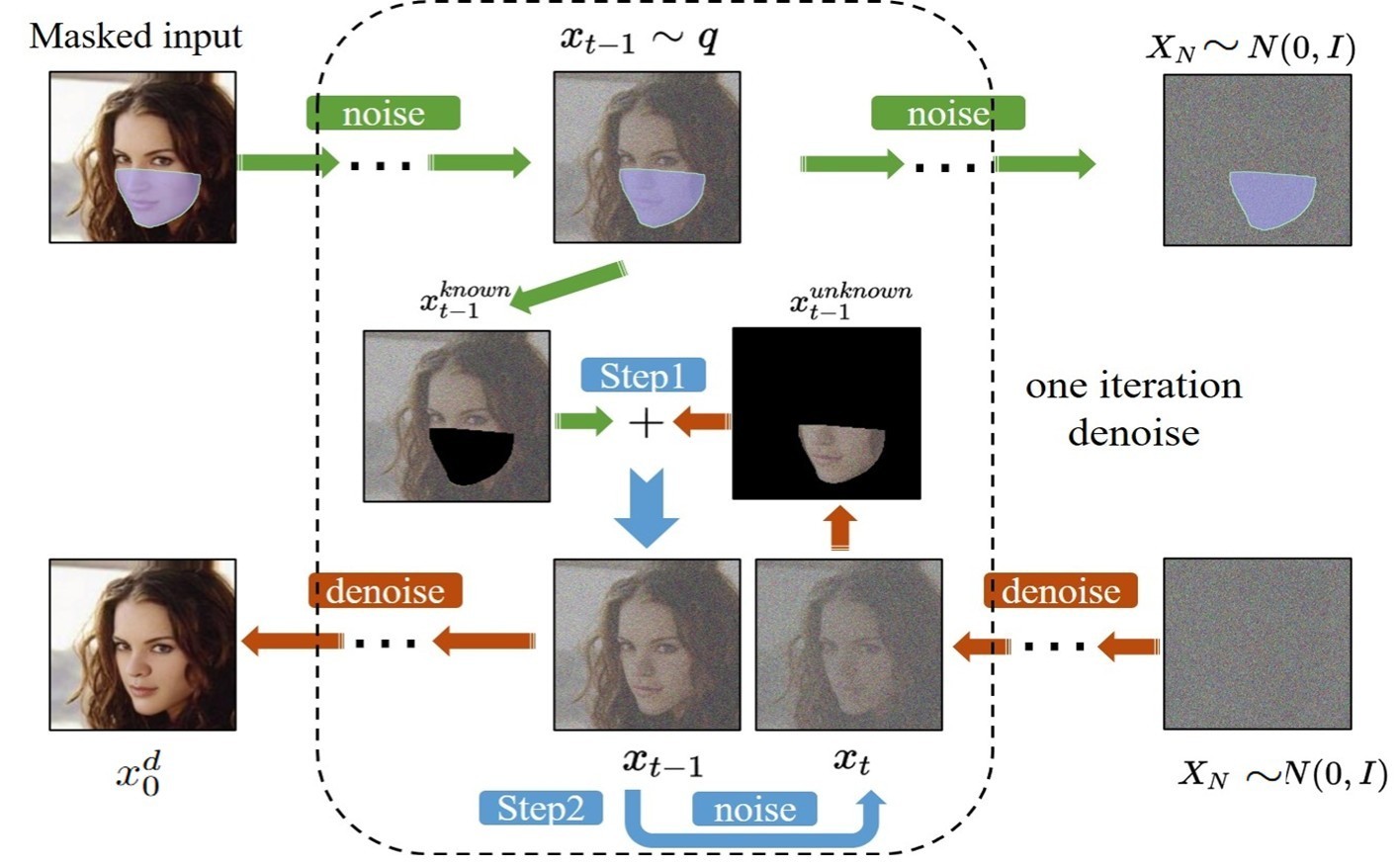}
\vspace{-1pt}
\caption{Face Repainting. Repaint modifies the standard denoising process to condition on the given image content. In each step, it samples the known region (top) from the input and the repainted part from the output (bottom).}
\label{fig2}
\vspace{-7px}
\end{figure}

As shown in Fig \ref{fig2}, the repainting model is improved in the denoising process by dividing $x_{t-1}$ into two parts: $m \odot x^{known}_{t-1}$, consisting of real images, and $(1-m) \odot x^{unknown}_{t-1}$, consisting of denoising results from $x_{t}$.
% \vspace{-4px}
\begin{equation}
\vspace{-3px}
x_{t-1}=m \odot x_{t-1}^{\textit{known}}+(1-m) \odot x_{t-1}^{\textit {unknown }},
\label{f1}
\end{equation}
\begin{equation}  
\vspace{-3px}
x_{t-1}^{\textit {known}}  \sim \mathcal{N}\left(\sqrt{\bar{\alpha}_{t}} x_{0}^{d},\left(1-\bar{\alpha}_{t}\right) \mathbf{I}\right),
\label{f2}
\end{equation}
\begin{equation}  
x_{t-1}^{\textit {unknown}}  \sim \mathcal{N}\left(\mu_{\theta}\left(x_{t}, t\right), \Sigma_{\theta}\left(x_{t}, t\right)\right),
\label{f3}
% \vspace{-6px}
\end{equation}
where $m$ is a mask matrix with elements in $[0,1]$,  $\odot$ denotes element-wise multiplication,  $x^{known}_{t-1}$ is sampled using the known pixels in the given images, while $x^{unknown}_{t-1}$ is sampled from the model, given the previous iteration $x_t$. This approach ensures that at each denoising iteration, $x_t$ is determined solely by $x_{t-1}$, and the distribution of the original image $x_0^{d}$ is incrementally incorporated in each iteration.

Although, it effectively enhances the quality of the output image, there will be a problem of boundary discontinuity during the stitching process. To solve this problem, a $j$ iterative resampling step is added to the $x_t$ and $x_{t-1}$ steps, so that the boundaries in the output image are aligned consistently. 
% This method has been shown to be effective ~\cite{lugmayr2022repaint}.

\setlength{\floatsep}{5pt plus 2pt minus 2pt}
\setlength{\textfloatsep}{5pt plus 2pt minus 2pt}
\setlength{\intextsep}{5pt plus 2pt minus 2pt}

The framework, illustrated in Fig. \ref{fig2}, involves an iterative denoising method for repainting an occluded face image. Each iteration consists of two steps: Firstly, we synthesize and splice $x_{t-1}$ by combining the known and unknown parts using formulas \ref{f1}, \ref{f2}, and \ref{f3}. Secondly, we resample $x_{t-1}$ by adding noise to obtain $x_{t}$. We perform the two steps $r$ times for one denoising iteration, which differs from previous work by incorporating resampling to improve the boundary, since face images have complex boundaries, and simple synthesis methods can create unnatural transitions and artifacts. By adding a resampling step and introducing noise, the model can better mimic the randomness in natural images, enhancing boundary smoothness.

\subsection{ID-Gated MoDE}

\subsubsection{Identity Gating Network}

To effectively assemble information from multiple reconstructed facial images, we introduce an identity gating network (ID-Gate) ~\cite{he2016identity} for evaluating the individual contributions of each reconstruction to the identity, and adaptively integrating the predicted results in the decision space. 
The ID-Gate takes a set of facial features extracted from the occluded image and repainted images as input, and produces a weight vector for each image in the decision space as output.

Specifically, for a sequence of facial images denoted as $X=[x_0,x_1,\dots,x_n]$, the extracted features are represented as $F(X)=[f(x_0),f(x_1),\dots,f(x_n)]$, which are then fed into the ID-Gate. 
The weight vector, denoted as $W(X)$, is utilized to integrate the predicted results of the facial images within the decision space, and is calculated as:
\begin{equation}
W(X) = softmax(W_{g} \cdot  F(X)+b_{g}),
\end{equation}
This computes the weights for fusing experts’ predictions via ID-Gate. where $W_{g}$ and $b_{g}$ are trainable parameters. 

It is imperative to highlight that the ID-Gate distinguishes itself from the gating network employed in the typical Mixture of Experts models on the following grounds:

\textbf{Network input/output:} 
Most gating networks take a single feature as input, making them a single-input network, and output weights for the features of different expert networks. However, Our ID-Gate receives features from multiple experts as inputs, making it a multi-input network, and produces weights for predictions. It endows each expert with specific knowledge for the corresponding face repainting diffusion model.

\textbf{Network objective:} The objective of most gating networks in the typical Mixture of Experts model aims to fuse features within the feature space. While our ID-Gate is designed to integrate identity prediction results based on deep representation features, thereby implicitly  comparing the reconstructed results with the original occluded faces for more effective, interpretable, and intuitive decision fusion.

\subsubsection{Mixture of Diffusion Experts}
Based on face repainting and the ID-Gate, we propose a novel identification framework, called Mixture of Diffusion Experts (MoDE). It consists of $n+1$ "experts", denoted as $E_0, E_1, \dots, E_n$, along with the ID-Gate, as depicted in Fig. \ref{fig1}. Within each expert, the repainted face and the feature extraction network are employed to generate the feature vectors of the experts. The feature extraction network within each expert shares parameters to ensure consistency. In order to retain the identity information of the image, we designate $E_0$'s image as the original image with occlusion, denoted as $x_0$. As the diffusion model serves as the foundation for face repainting in our work, each "expert" is taken as a diffusion expert.

Given the input $x_0$, the repaint process generates a sequence of images for each expert, denoted as $X=[x_0, x_1, \dots, x_n]$. Subsequently, the feature extraction network is employed to obtain the feature output $f(x_i)$ of the experts. The similarity matrix $s(x_i)$, calculated based on the features, is used to represent the prediction results of the faces in the gallery, and the features are also fed into the ID-Gate to generate the weight vector $w(x_i)$ for integrating the predictions in the decision space. Finally, the Mixture step is performed, where the results of the diffusion experts are integrated using the weight vector.
\begin{equation}
\vspace{-6px}
S_{X} = \sum_{i=0}^{n} w(x_i) s(x_i).
\end{equation}

By applying the argmax function on the weighted similarity matrix $S_X$, the best matching face can be identified. 
% Hence, we term the network structure that combines ID-Gate and Diffusion Experts as Mixture of Diffusion Experts (MoDE).

\subsection{Objective Loss}
In MoDE, the overall objective loss is decomposed into two components: the loss incurred during the pre-training of the diffusion model and the training loss of ID-Gate. During the training of the diffusion model, the loss function is optimized through the minimization of the negative log-likelihood of the model in training set, which can be mathematically expressed as:
\begin{equation}  
\mathcal{L}_{simple }=E_{t, x_{0}^{d}, \epsilon}\left[\left\|\epsilon-\epsilon_{\theta}\left(x_{t}, t\right)\right\|^{2}\right],
\end{equation}
where $\mathcal{L}_{simple}$ represents the loss function, $\epsilon_{\theta}\left(x_{t}, t\right)$ represents a learned denoising diffusion process, and $\left\| \cdot \right\|$ represents a Euclidean distance metric. The expectation over $t$, $x_{0}^{d}$, $\epsilon$ ensures robust denoising of occluded inputs under varying timesteps and noise. Moreover, $\mu_{\theta}\left(x_{t}, t\right)$ defines the mean of the latent variables of the diffusion model as:
\begin{equation}  
\mu_{\theta}\left(x_{t}, t\right)=\frac{1}{\sqrt{\alpha_{t}}}\left(x_{t}-\frac{\beta_{t}}{\sqrt{1-\bar{\alpha}_{t}}} \epsilon_{\theta}\left(x_{t}, t\right)\right).
\end{equation}

The ID-Gate is optimized by the cross-entropy loss:
\begin{equation}
\begin{split}
\mathcal{L} &=\frac{1}{B}\sum_{k=1}^{B} l(S_{X_{k}},label_{k})\\ &= -\frac{1}{B}\sum_{k=1}^{B} log(\frac{\exp(S_{X_{k}}[label_{k}])}{\sum_{i=1}^{M}\exp(S_{X_{k}}[i])}),
\end{split}
\end{equation}
where $B$ and $S_{X_{k}}$ are the batch size and weighted similarity matrix of the $k$-th identity, $M$ denotes the total number of identities, and $label_{k}$ is the label of the k-th identity.

\subsection{Network Architecture}
The ID-Gate in MoDE can be one of the following two network architectures.

\textbf{Softmax Gating:} A straightforward choice of gating network~\cite{jordan1994hierarchical} involves multiplying the input by a trainable weight matrix $W_{g}$, adding the bias $b_{g}$, and then applying the Softmax function. 
% \begin{equation}
%     % G(x)=\operatorname{Softmax}\left( W_{g} \cdot x + b_{g}\right)
%     G(x)={Softmax}\left( W_{g} \cdot x + b_{g}\right).
% \end{equation}

\vspace{-1px}
\textbf{Noisy Top-K Gating:} The Softmax gating network can be augmented with two components: sparsity~\cite{shazeer2017outrageously} and noise. Before applying the Softmax function, tunable Gaussian noise is injected through the ${StandardNormal()}$ term which adds i.i.d. Gaussian noise $\epsilon_i \sim \mathcal{N}(0, 1)$ into each gating component, encouraging exploration during training. Only the top k values are retained, with the rest set to $-\infty$. The sparsity improves computation efficiency. The amount of noise per component is controlled by a second trainable weight matrix $W_{noise}$,
\begin{equation}
\small
    % G(x)=\operatorname{Softmax}(KeepTopK(H(x), k))
    G(x)={Softmax}(K-TopK(H(x), k)),
\end{equation}
\begin{equation}
\hspace{-0em}
\footnotesize
H(x)_{i}=\left(x \cdot W_{g}\right)_{i}+StandardNormal() \cdot Softplus\left(\left( W_{noise }  \cdot x \right)_{i}\right),
\end{equation}
\begin{equation}
\hspace{-0em}
\small
    K-TopK (v, k)_{i}=\left\{\begin{array}{ll}
v_{i} & \text { if } v_{i} \text { is in the top } k \text { elements of } v . \\
-\infty & \text { otherwise. }
\end{array}\right.
\end{equation}

For ID-Gate, we use the first Softmax Gating, and the second Noisy Top-K Gating is discussed in the appendix.

\section{Experiments}

% In this section, we present the experimental settings, including the dataset, evaluation metrics and implementation details. We then design three experiments to validate the superiority of our proposed MoDE method in face recognition tasks. 
We first compare our method with five state-of-the-art face recognition methods, including ArcFace~\cite{deng2019arcface}, FaceNet~\cite{schroff2015facenet,cao2018vggface2}, CosFace~\cite{wang2018cosface}, FFR-Net~\cite{hao2022unified}, and Deepface-EMD~\cite{phan2022deepface}. 
Then, we validate the MoDE on two real datasets collected by ourselves: the Occluded Volunteer Face (OVF) dataset and the Web Wild Occluded Celebrity Face (WWCF) dataset. 
% These datasets contain images of faces in complex environments, including occlusions and varying lighting conditions. Furthermore, we perform ablation studies to analyze the contributions of different modules in our MoDE. 
% In addition, We also provide more experimental results in anonymous link \url{https://anonymous.4open.science/r/MODE-Supplementary-Material-5ACB}.  

% \begin{table}[t]
% \caption{The five state-of-the-art comparisons models.}\label{tab44}
% \vspace{-1pt}
% \centering
% \vskip 0.1mm \setlength{\tabcolsep}{0.7mm}
% \begin{tabular}{@{\extracolsep\fill}cccc}
% \toprule
% Model & Architectures & Train Dataset & Year \\
% \midrule
 % FaceNet ~\cite{schroff2015facenet}
 % & InceptionResnetV1 & VGGFACE2 & 2018   \\ 
% CosFace ~\cite{wang2018cosface} & 20-layersphereface & CAISA-WebFace  & 2018 \\
% ArcFace~\cite{deng2019arcface} & ResNet-50 & MS1M  & 2019  \\
% FFR-Net ~\cite{hao2022unified}& SENet50  & CAISA-WebFace  & 2022 \\
% DeepFace-EMD~\cite{phan2022deepface} & ResNet-18  & CAISA-WebFace  & 2022 \\
% \bottomrule
% \end{tabular}
% \end{table}

% \setlength{\floatsep}{3pt plus 2pt minus 2pt}
% \setlength{\textfloatsep}{3pt plus 2pt minus 2pt}
% \setlength{\intextsep}{3pt plus 2pt minus 2pt}
\vspace{-1px}
\subsection{Experimental Setting}

\noindent\textbf{Datasets:} We conducted experiments on three datasets: 
MS1M~\cite{deng2019arcface}, LFW~\cite{huang2008labeled} and CelebA~\cite{liu2015deep}. MS1M consists of over one million face images from 13,126 different identities, LFW consists of 13,233 face images from 5,749 identities, while CelebA contains 202,599 face images from 10,177 different identities. 

To simulate mask occlusion scenarios, we used facial keypoint detection to add mask textures to the face images in these datasets, creating three new sub-datasets: Occ MS1M, Occ LFW, and Occ CelebA. During the training phase, we selected 500 faces in the Occ MS1M dataset as the probe, and 2500 faces in MS1M as the gallery. During the testing phase, we randomly selected 500 faces in Occ MS1M, 1000 faces in Occ LFW and 1000 faces in Occ CelebA as the probe, respectively. We used 2500 faces in MS1M, 4000 faces in LFW and 5000 faces in CelebA as the gallery, respectively. 
%In addition, we also randomly select 2000 pairs of images on each of the above three datasets for face verification.

\noindent\textbf{Evaluation Metrics:} Our MoDE model was evaluated using four widely used evaluation metrics: $Top1$, $Top5$, $EER$, and $Acc$. $Top1$ and $Top5$ refer to the Top-1 and Top-5 face recognition accuracy rates respectively. For face verification, $EER$ (Equal Error Rate) is a well-established metric in biometric verification systems and represents the point at which the $FAR$ (False Acceptance Rate) and $FRR$ (False Rejection Rate) are equal. $Acc$ indicates the model's  verification accuracy at the $EER$ threshold.

\noindent\textbf{Implementation Details:} 
For image preprocessing, we first performed face detection on the images and then cropped and resized them to 112$\times$112 pixels. To prevent overfitting and improve the generalization ability of our model, we applied data augmentation when training the Gate network. The input of the Gate network was a feature vector concatenated from $n+1$ feature vectors.
% which had $(n+1)!$ possible permutations, allowing us to augment our training set by $(n+1)!$ times. 
We used the Adam optimizer with a learning rate of 1e-6, a batch size of 32, and trained for 200 epochs. 
% As using the diffusion model for image generation required high computational resources, we used a GPU with an RTX 3090 (32GB memory).
For the repaint model, we set time steps $T = 200$, resampling times $r = 10$, and jumpy size $j = 10$.  
% These hyperparameters were chosen based on empirical experimentation. The differences between our network architecture and other state-of-the-art methods are obviously.
 % shown in Table~\ref{tab44}. 
In the ablation study, we adopted the ResNet-50 convolutional neural network used by ArcFace~\cite{deng2019arcface} and pretrained on the MS1M dataset for feature extraction.

\noindent\textbf{Competing Methods:} ArcFace~\cite{deng2019arcface} is a deep face recognition method that improves the discriminability of face features by adding an angular margin to the feature space. It enhances the intra-class compactness and inter-class differences by increasing the normalization of the feature vector. FaceNet~\cite{schroff2015facenet} is a convolutional neural network-based models that maps face images to a low-dimensional vector space, known as face embedding. It determines whether two face images belong to the same person by comparing their embedding vectors. CosFace~\cite{wang2018cosface} uses a margin-based softmax loss function similar to Sphereface~\cite{liu2017sphereface} to constrain the weight matrix $W$ in the classifier to a hypersphere, resulting in greater cosine similarity of face features for the same person and less cosine similarity between different people. FFR-Net~\cite{hao2022unified} uses an encoder-decoder structure to map the occluded region features to the unoccluded region, thus improving the accuracy of face recognition. Deepface-EMD~\cite{phan2022deepface} is a re-ranking approach that compares two faces using the Earth Mover’s Distance on the deep, spatial features of image patches.

 % For image preprocessing, we first performed face detection on the images, and cropped and resized them to 112*112 pixels. To prevent overfitting and improve the generalization ability of our model, we applied data augmentation when training the Gate network. The input of the Gate network was a feature vector concatenated from $n+1$ feature vectors, which had $(n+1)!$ possible permutations. Therefore, we could augment our training set by $(n+1)!$ times. We used Adam optimizer with learning rate=1e-6, batch$\_$size=32, num$\_$epoch=200 for training. Since using diffusion model for image generation required high computational resources, we used GPU with RTX 3090 (32GB memory).

\vspace{-3px}

\subsection{Comparisons}
To evaluate the effectiveness of our proposed method (MoDE), in comparison with existing mainstream face recognition techniques under occlusion, we evaluated five competing algorithms on three artificial occluded datasets: MS1M, LFW, and CelebA, and two real occluded datasets: OVF and WWCF.

\subsubsection{Results on Artificial Occluded Face}
%To compare the effectiveness of our proposed method (MoDE) with mainstream face recognition algorithms in the presence of occlusion, we conducted experiments on three Masked datasets, MS1M, LFW and CelebA, and tested six different algorithms: ArcFace, FaceNet, CosFace, FFR-Net, Deepface-EMD, and our MoDE method.

%We evaluated the performance of these algorithms under occlusion by comparing their recognition rates with different occlusion ratios. Our MoDE method achieved the highest accuracy on both datasets, demonstrating its superiority over the other algorithms.

\begin{table}[t]
\caption{Comparison between state-of-the-art models and MoDE in face recognition tasks on occluded MS1M, LFW and CelebA datasets. Our MoDE outperforms all other methods.
% Our MoDE method has improved accuracy over the other six methods, with a maximum improvement of 13.7$\%$ for Top1 (DeepFace: 22.2$\%$-35.9$\%$) and 15$\%$ for Top5 (DeepFace: 40.2$\%$-55.2$\%$).
}\label{tab66}
\vspace{-3pt}
\centering
\vskip 0.1mm \setlength{\tabcolsep}{1.5mm}
\resizebox{.99\columnwidth}{!}{
\begin{tabular}{@{\extracolsep\fill}cccll}
\toprule
Dataset & Model & Method & Top1 & Top5 \\ 
\midrule
\multirow{14}{*}{\makecell[c]{Occ \\ MS1M}} & \multirow{2}{*}{ArcFace} & baseline & 87.8 & 93.6\\
~ & ~ & MoDE & 88.8(\textcolor{red}{+1.0}) & 95.0(\textcolor{red}{+1.4})\\
\cmidrule{2-5}
~ & \multirow{2}{*}{FaceNet} & baseline & 30.6 & 47.4\\
~ & ~ & MoDE & 31.6(\textcolor{red}{+0.2}) & 49.0(\textcolor{red}{+1.6})\\
\cmidrule{2-5}
~ & \multirow{2}{*}{CosFace} & baseline & 42.4 & 59.0\\
~ & ~ & MoDE & 47.2(\textcolor{red}{+4.8}) & 63.2(\textcolor{red}{+4.2})\\
\cmidrule{2-5}
~ & \multirow{2}{*}{FFR-Net} & baseline & 70.6 & 83.4\\
~ & ~ & MoDE & 74.2(\textcolor{red}{+3.6}) & 84.2(\textcolor{red}{+0.8})\\
\cmidrule{2-5}
~ & \multirow{4}{*}{DeepFace-EMD} & baseline & 18.6 & 32.6\\
~ & ~ & EMD & 25.6(\textcolor{red}{+7.0}) & 39.0(\textcolor{red}{+6.4})\\
~ & ~ & MoDE & 31.2(\textcolor{red}{+12.6}) & 48.8(\textcolor{red}{+16.2})\\
~ & ~ & EMD+MoDE & 38.2(\textcolor{red}{+19.6}) & 54.3(\textcolor{red}{+21.7})\\
\midrule
\multirow{14}{*}{\makecell[c]{Occ \\ LFW}} & \multirow{2}{*}{ArcFace} & baseline & 64.6 & 79.3\\
~ & ~ & MoDE & 71.4(\textcolor{red}{+6.8}) & 82.0(\textcolor{red}{+2.7})\\
\cmidrule{2-5}
~ & \multirow{2}{*}{FaceNet} & baseline & 27.6 & 51.4\\
~ & ~ & MoDE & 34.2(\textcolor{red}{+6.6}) & 56.7(\textcolor{red}{+5.3})\\
\cmidrule{2-5}
~ & \multirow{2}{*}{CosFace} & baseline & 10.4 & 18.4\\
~ & ~ & MoDE & 14.4(\textcolor{red}{+4.0}) & 24.0(\textcolor{red}{+5.6})\\
\cmidrule{2-5}
~ & \multirow{2}{*}{FFR-Net} & baseline & 46.7 & 62.6\\
~ & ~ & MoDE & 52.5(\textcolor{red}{+5.8}) & 67.6(\textcolor{red}{+5.0})\\
\cmidrule{2-5}
~ & \multirow{4}{*}{DeepFace-EMD} & baseline & 22.2 & 40.2\\
~ & ~ & EMD & 30.2(\textcolor{red}{+8.0}) & 48.5(\textcolor{red}{+8.3})\\
~ & ~ & MoDE & 35.9(\textcolor{red}{+13.7}) & 55.2(\textcolor{red}{+15})\\
~ & ~ & EMD+MoDE & 47.5(\textcolor{red}{+25.3}) & 63.9(\textcolor{red}{+23.7})\\
\midrule
\multirow{14}{*}{\makecell[c]{Occ \\ CelebA}} & \multirow{2}{*}{ArcFace} & baseline & 23.2 & 38.4\\
~ & ~ & MoDE & 31.9(\textcolor{red}{+8.7}) & 46.0(\textcolor{red}{+7.6})\\
\cmidrule{2-5}
~ & \multirow{2}{*}{FaceNet} & baseline & 16.4 & 31.0\\
~ & ~ & MoDE & 21.5(\textcolor{red}{+5.1}) & 37.4(\textcolor{red}{+6.4})\\
\cmidrule{2-5}
~ & \multirow{2}{*}{CosFace} & baseline & 3.4 & 9.7\\
~ & ~ & MoDE & 9.6(\textcolor{red}{+6.2}) & 17.0(\textcolor{red}{+7.3})\\
\cmidrule{2-5}
~ & \multirow{2}{*}{FFR-Net} & baseline & 15.0 & 26.0\\
~ & ~ & MoDE & 20.2(\textcolor{red}{+5.2}) & 32.5(\textcolor{red}{+6.5})\\
\cmidrule{2-5}
~ & \multirow{4}{*}{DeepFace-EMD} & baseline & 24.6 & 42.0\\
~ & ~ & EMD & 32.2(\textcolor{red}{+7.6}) & 48.0(\textcolor{red}{+6.0})\\
~ & ~ & MoDE & 31.6(\textcolor{red}{+7.0}) & 50.2(\textcolor{red}{+8.2})\\
~ & ~ & EMD+MoDE & 37.3(\textcolor{red}{+12.7}) & 54.3(\textcolor{red}{+12.3})\\
\toprule
\end{tabular}
}
\vspace{-3pt}
\end{table}

% In our study, we compares the performance of five face recognition algorithms, including ArcFace, FaceNet, CosFace, FFR-Net and Deepface-EMD, with our MoDE method on three test sets. Deepface-EMD and MoDE are similar in that they are both plug-and-play modules. Therefore, for our comparative experiments on Deepface-EMD, we made the following additional settings: EMD represents  the use of Deepface-EMD, MoDE represents  the use of MoDE, and EMD+MoDE represents  the simultaneous use of both methods. Our results, as shown in Table \ref{tab66}, indicate that MoDE outperforms all other methods in terms of recognition accuracy, regardless of the feature extraction network used. In particular, we observed a substantial boost in the accuracy of the Maked LFW and CelebA datasets compared to the Occluded MS1M dataset.
% In the face verification task, the MoDE method outperforms other methods in terms of Acc, TPR, EER, and AUC metrics, as shown in Table \ref{tab666}.

Table \ref{tab66} shows that MoDE outperforms all the competing methods in terms of recognition accuracy. Notably, we observed a significant increase in accuracy on the Occ LFW and Occ CelebA datasets compared to the Occ MS1M dataset.

FaceNet, ArcFace, and CosFace are feature extractors based on metric learning, aiming to increase intra-class compactness and inter-class separability. However, due to the absence of large-scale occluded face data for training, they are still affected by occlusion, resulting in a decline in recognition accuracy. FFR-Net is a method that focuses on converting occluded face features into unoccluded face features. However, its performance in real-world scenarios is not ideal, and it may not be capable of handling occluded faces well. Deepface-EMD uses Earth Mover’s Distance to add additional comparison stages to explicitly check image similarity at a fine-grained level. Similarly, MoDE is also a plug-and-play module that can be widely applied to any traditional face recognition model, achieving higher recognition accuracy than Deepface-EMD on Occ MS1M and Occ LFW. Moreover, we find the largest improvement when using both MoDE and Deepface-EMD, indicating that the two modules do not conflict.

Our proposed MoDE method addresses the occlusion problem by using a repaint model to generate high-quality reconstructions of occluded faces, and leveraging the ID-Gate module to selectively use input images with reliable identity information. Jointly using original and reconstructed image information produces superior performance in recognizing occluded faces. Hence, MoDE has inherent advantages in dealing with occlusion and could be considered a promising approach for occluded face recognition.

\subsubsection{Results on Real Occluded Face}

\begin{table}[t]
\caption{Comparisons in real scene.
% Bold \textcolor{red}{red} indicates the best. Bold \textcolor{blue}{blue} indicates the second best.
}\label{tab4}
\vspace{-3pt}
\centering
\vskip 0.1mm \setlength{\tabcolsep}{1mm}
\resizebox{.95\columnwidth}{!}{
\begin{tabular}{@{\extracolsep\fill}cccccc}
\toprule
\multirow{2}{*}{Model} & \multirow{2}{*}{Method} & \multicolumn{2}{@{}c@{}}{OVF} &  \multicolumn{2}{@{}c@{}}{WWCF}\\ 
\cmidrule{3-4}\cmidrule{5-6}
~ & ~ & Top1 & Top5 & Top1 & Top5\\
\midrule
\multirow{2}{*}{ArcFace} & baseline & 89.8 & \textcolor{red}{100} & 22.0 & 36.7\\
~ & MoDE & \textcolor{red}{95.5} & \textcolor{red}{100} & \textcolor{red}{22.5} & \textcolor{red}{37.0}\\
\midrule
\multirow{2}{*}{FaceNet} & baseline & 64.8 & \textcolor{red}{86.4} & \textcolor{red}{8.6} & 21.3\\
~ & MoDE & \textcolor{red}{65.9} & \textcolor{red}{86.4} & 7.6 & \textcolor{red}{22.0}\\
\midrule
\multirow{2}{*}{CosFace} & baseline & 83.0 & 96.6 & 4.1 & 8.7\\
~ & MoDE & \textcolor{red}{93.2} & \textcolor{red}{100} & \textcolor{red}{5.3} & \textcolor{red}{10.8}\\
\midrule
\multirow{2}{*}{FFR-Net} & baseline & 96.6 & \textcolor{red}{100} & 9.6 & 22.3\\
~ & MoDE & \textcolor{red}{98.9} & \textcolor{red}{100} & \textcolor{red}{12.3} & \textcolor{red}{25.3}\\
\midrule
\multirow{4}{*}{Deepface-EMD} & baseline & 69.3 & \textcolor{blue}{93.2} & 16.4 & 31.7\\
~ & EMD & \textcolor{blue}{90.9} & \textcolor{red}{98.9} & 19.7 & 34.4\\
~ & MoDE & {87.5} & \textcolor{red}{98.9} & \textcolor{blue}{21.2} & \textcolor{blue}{39.4}\\
~ & EMD + MoDE & \textcolor{red}{95.5} & \textcolor{red}{98.9} & \textcolor{red}{24.9} & \textcolor{red}{41.6}\\
\toprule
\end{tabular}
}
\vspace{-6pt}
\end{table}

% To evaluate the effectiveness of our proposed method in addressing occlusion challenges in real-world scenarios, we collected two occluded face datasets and conducted experiments on them. 
We evaluated the effectiveness of our proposed method in addressing occlusion challenges in real-world scenarios by collecting two occluded-face datasets and conducting experiments on them. The first, called the Occluded Volunteer Face Dataset (OVF), was obtained from volunteers wearing masks captured in various public settings, containing 352 face images of 44 identities. The second one, called the Web Wild Occluded Celebrity Face (WWCF) dataset, was collected from publicly available web images, including 873 face images from 233 identities.
% Mask occlusions pose a common challenge in face recognition, as they significantly impact face feature extraction and matching. 
% We conducted experiments with five state-of-the-art methods, and the results are summarized in Table~\ref{tab4}.

As shown in Table \ref{tab4}, MoDE performs better in dealing with face occlusion problems compared with most face recognition models. 
% , demonstrating strong generalization ability and robustness
In contrast to artificial occluded datasets, real occluded datasets present greater challenges in face recognition tasks, due to the greater variability in occlusion location and degree, as well as the unpredictability of occlusion quality and shape. Fortunately, the repaint process of MoDE effectively resolves these issues by restoring features in occluded areas to the greatest extent possible. The ID-Gate module's adaptive selection mechanism guarantees the quality and credibility of input data, enabling more efficient utilization of identity information. Moreover, it is noteworthy that MoDE has demonstrated remarkable performance on small-scale datasets when incorporated with specific methodologies (FFR-Net, OVF, 98.9\%), indicating its potential practical utility.

% This is attributed to the adaptive selection mechanism of the ID-Gate module for the quality and credibility of the input data.

% \begin{figure}[t]
% \centering
% \includegraphics[width=0.47\textwidth]{figure/Figure5.png}
% \caption{The dataset on the left is the high quality dataset we shot, and the dataset on the right is the one we made on the web. Each identity has one unmasked face and two faces wearing different masks.}
% \label{fig5}
% \end{figure}
\vspace{-2px}

\subsection{Ablation Study}

%To evaluate the effectiveness of our proposed method, we conducted ablation studies to analyze the contribution of different modules and compare experimental results under various settings. We designed several variants of our method to establish baselines and test the impact of key components. The variants include:
% To assess the efficacy of MoDE, we performed ablation studies aimed at scrutinizing the contribution of individual modules and comparing experimental results across different configurations.
We performed ablation studies as follows:
% These experiments aimed to scrutinize the contribution of individual modules and compare the experimental results across different configurations. 

% to establish baselines and evaluate the influence of key components. The variants include:

\begin{itemize}
\item [(1)] \textbf{Baseline}: ResNet-50 was served as the feature extractor for occluded face recognition directly.
\item [(2)] 
\textbf{Baseline + RF}: Diffusion experts were introduced to generate multiple repainted faces (RF), which were subsequently fused by averaging.
\item [(3)] 
\textbf{MoDE}: We utilized ID-Gate to adaptively integrate the predictions of occluded faces and multiple repainted faces in the decision space.
\end{itemize}

\begin{figure}[t]
\centering
\includegraphics[width=0.5\textwidth]{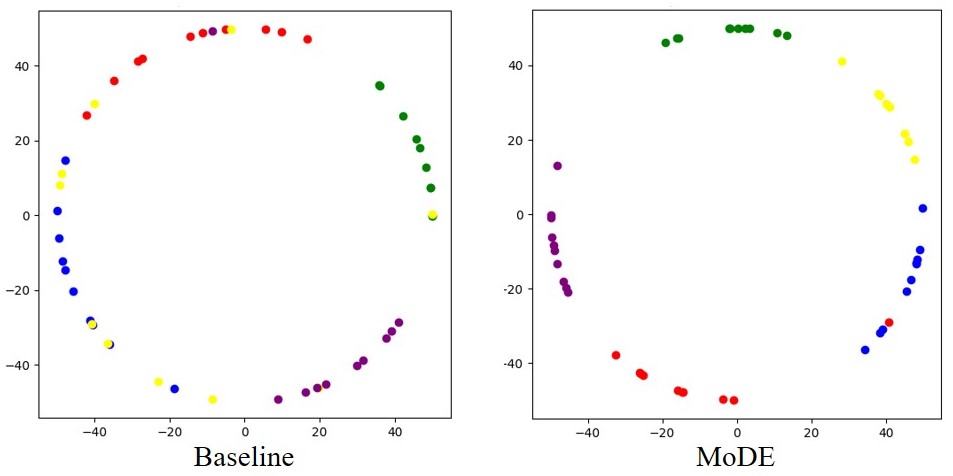}
\vspace{-8pt}
\caption{Visualization of similarity distributions with t-SNE~\cite{van2008visualizing}  and following normalization. Different markers with colors represent different classes.}
\label{fig6}
\end{figure}

\begin{table}[!t]
\caption{Comparisons of three variants on Occ CelebA.
}\label{tab11}
\vspace{-3pt}
\centering
\vskip 0.1mm \setlength{\tabcolsep}{1mm}
\resizebox{\linewidth}{!}{
\begin{tabular}{@{\extracolsep\fill}lllll}
\toprule%
Method  &Top1 & Top5 &Acc  & EER \\
\midrule
Baseline   & {23.2} & {38.4} & {83.5}  & {0.165} \\
Baseline + RF   & {28.7(\textcolor{red}{+5.5})}  & {41.1(\textcolor{red}{+2.7})} & {85.6(\textcolor{red}{+2.1})}  & {0.145(\textcolor{red}{-0.020})} \\
MoDE  & {31.9(\textcolor{red}{+8.7})}  & {46.0(\textcolor{red}{+7.6})} & {85.4(\textcolor{red}{+1.9})}  & {0.146(\textcolor{red}{-0.019})} \\ 
\toprule
\end{tabular}
}
\vspace{-3pt}
\end{table}

\setlength{\floatsep}{5pt plus 2pt minus 2pt}
\setlength{\textfloatsep}{5pt plus 2pt minus 2pt}
\setlength{\intextsep}{5pt plus 2pt minus 2pt}

% \noindent\textbf{Quantitative Results: }Table \ref{tab11} presents the  results of various experimental variants on the Occluded MS1M and CelebA datasets. Among all the variants, the MoDE model achieves the highest recognition top-1 and top-5 accuracy, with a 17.2$\%$ and 12.2$\%$ improvement over the Baseline top-1 and top-5 accuracy on the Occluded MS1M, respectively. Notably, removing either occluded or repainted faces leads to an increase in error rate, suggesting that both types of information are interdependent and essential. This finding further validates the effectiveness of our proposed method. In addition, when ID-Gate was directly added onto the Baseline approach, its effectiveness did not show significant improvement. This might be attributed to the fact that our diffusion experts shared a common generative model, thereby having an equal distribution of identity information among the repainted faces. The use of ID-Gate for decision-level fusion should assign an equal weight to each repainted face when sufficient samples are available, resembling average fusion.

\noindent\textbf{Quantitative Results: }Table \ref{tab11} presents the  results of various experimental variants on the Occ CelebA dataset. Among all the variants, the MoDE model achieved the highest recognition Top-1 and Top-5 accuracy, with an 8.7$\%$ and 7.6$\%$ improvement over the "Baseline" Top-1 and Top-5 accuracy, respectively. Leveraging the diverse identity information obtained from repainted images, the "Baseline + RF" approach achieved a certain level of improvement. However, it viewed all repainted images as equal and ignored the diversity of the diffusion model, failing to adaptively differentiate between relevant and redundant information for identification. In contrast, the incorporation of ID-Gate allowed for dynamic decision space integration of multiple experts to filter out noise from repainted images that failed to meet our expectations, thereby enhancing the model's adaptability and fault tolerance.

\begin{figure}[t]
\centering
\includegraphics[width=0.47\textwidth]{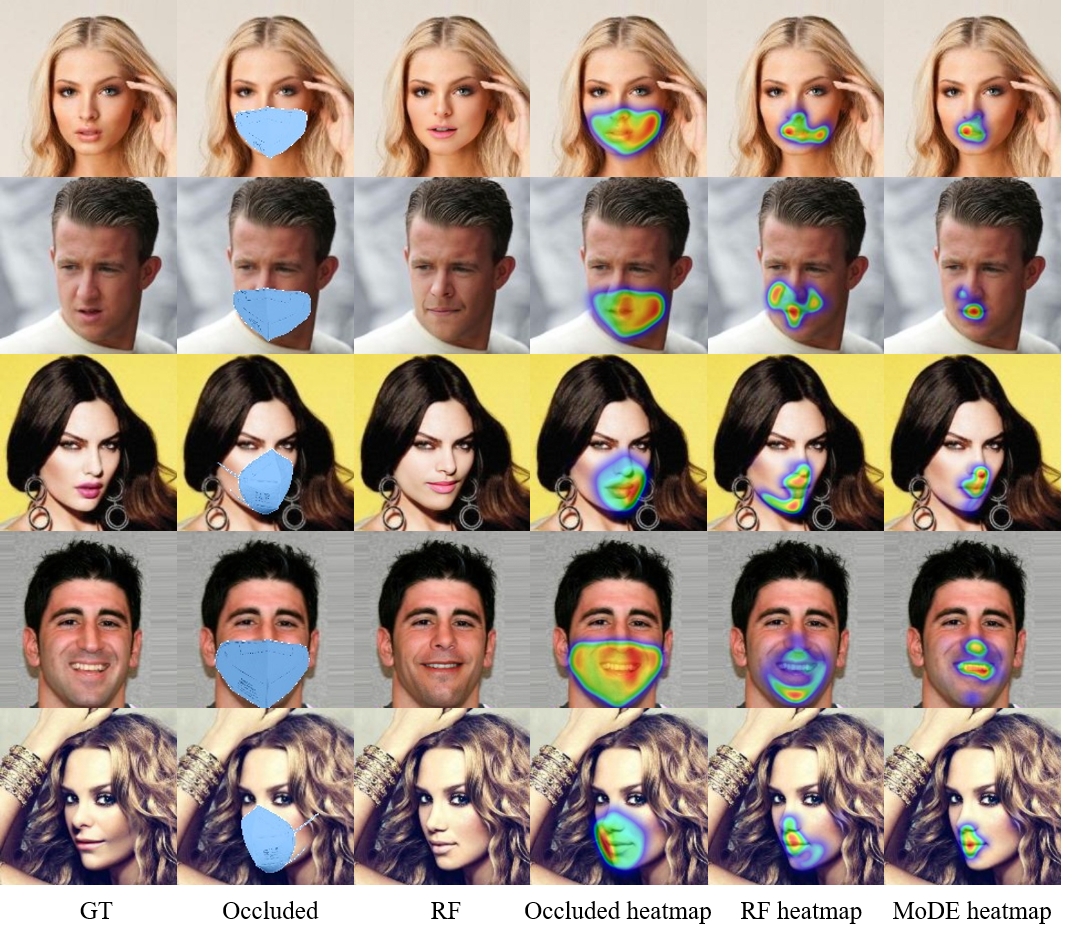}
\vspace{-3pt}
\caption{Visualization of face repaint error. The first three columns represent normal faces, occluded faces and repainted faces, respectively. The last three columns represent heatmaps indicating the degree of information loss between GT and Occluded, RF, and MoDE, respectively.}
\vspace{-3pt}
\label{fig3}
\end{figure}

\setlength{\floatsep}{2pt plus 2pt minus 2pt}
\setlength{\textfloatsep}{2pt plus 2pt minus 2pt}
\setlength{\intextsep}{2pt plus 2pt minus 2pt}

% \begin{figure}[t]
% \centering
% \includegraphics[width=0.47\textwidth]{figure/Figure33.png}
% \vspace{-15pt}
% \caption{Visualization of face repaint error. The former three columns represent normal faces, occluded faces and repainted faces, respectively. The latter three columns represent heatmaps indicating the degree of information loss between GT and Occluded, RF, MoDE, respectively.}
% \label{fig3}
% \end{figure}

% \setlength{\floatsep}{2pt plus 2pt minus 2pt}
% \setlength{\textfloatsep}{2pt plus 2pt minus 2pt}
% \setlength{\intextsep}{2pt plus 2pt minus 2pt}

\noindent\textbf{Qualitative Analyses: }To evaluate the effectiveness of MoDE in occluded face classification, we compared the similarity distributions of the baseline and MoDE on a subset of the Occ MS1M dataset. We randomly selected five classes, each with 10 samples, and used a gallery with 500 identities. Then we obtained the 500-dimensional similarity distribution for each model and transformed the features into 2D using t-SNE. The normalized results are depicted in Fig. \ref{fig6}. 
The similarity distribution of the baseline model is scattered, indicative of subpar performance in occluded face classification. In contrast, MoDE's similarity distribution is more compact and distinguishable, highlighting its outstanding capability to address occlusion challenges. This is credited  to its deep representation features obtained through decision space integration.

To investigate the benefits of the MoDE method in detail, we compared the heat maps of information loss identified when using occluded images, repainted images, and the MoDE method. As illustrated in Fig. \ref{fig3}, a larger area and more intense color on the heatmap indicate a more substantial loss of information. Occluded images result in complete loss of information in the occluded regions (as seen by the area and color of the heat map). Repainted images may alleviate the problem, but significant loss still remains in some regions. In contrast, the MoDE method yielded the best results, emphasizing its superior capability in both minimizing information loss and maximizing identity utilization for recognition.

% \begin{figure}[h]
% \centering
% \includegraphics[width=0.5\textwidth]{figure/Figure4.jpg}
% \caption{ (a) and (b) report CMC curve and ROC curve of  Recognition accuracy($\%$) with different models. }
% \label{fig4}
% \end{figure}

% Figure \ref{fig4} shows the Receiver Operating Characteristic (ROC) and Cumulative Match Characteristic (CMC) curves. The curves confirm that the MoDE model outperforms the Baseline model and the DDPM model in terms of True Acceptance Rate (TAR) at various False Acceptance Rate (FAR) levels. This demonstrates that the use of the DDPM model can effectively restore the occluded regions and improve the feature representation capability. Furthermore, the use of the ID-Gate module can efficiently exploit the complementary information between the original and reconstructed images, resulting in enhanced discriminative ability.
\vspace{-4px}
\subsection{Discussion}

\begin{figure}[t]
\centering
\includegraphics[width=0.49\textwidth]{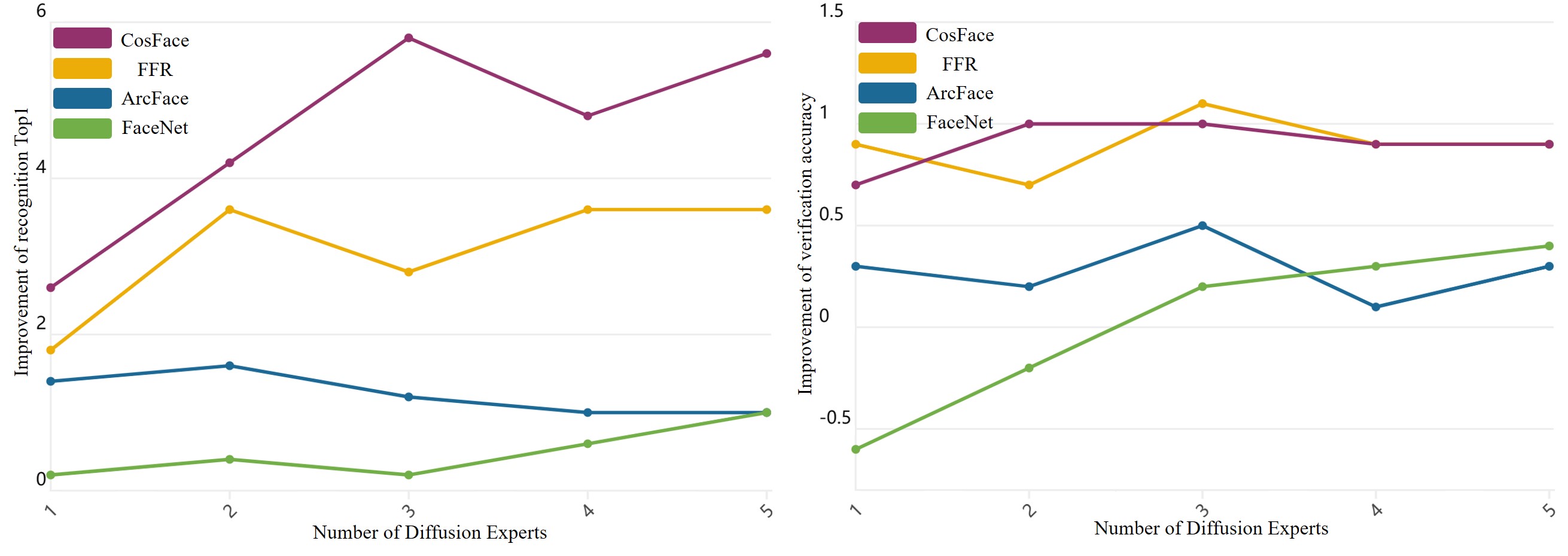}
% \vspace{-6pt}
\caption{Hyperparameter analysis of the MoDE method with varying numbers of Diffusion experts. }
\label{fig7}
\end{figure}

\noindent\textbf{Hyperparameter Analysis: }We conducted a hyperparameter analysis to determine the optimal value of $n$ that indicates the number of diffusion experts in the MoDE framework. The analysis was performed on the MS1M dataset, focusing on identification and verification tasks. Two graphs presented in Fig. \ref{fig7} illustrates the improvement of Top-1 recognition ($\%$) and verification accuracy (Acc) ($\%$). Our findings demonstrate that as the number of experts defined by $n$ increases, there is a consistent upward trend in both the Top-1 and Acc metrics. The improvement appears to stabilize upon reaching a threshold of four to five experts.

% We visually present the results in Figure \ref{fig7}. To evaluate the MoDE framework, we utilized four well-known baseline algorithms, ArcFace, FFR-Net, CosFace, and FaceNet.Notably, CosFace and FFR-Net algorithms demonstrated the most significant improvement within the MoDE framework.  Furthermore, we observed a consistent increasing trend in the improvement of Top1 and Acc metrics with an increase in the number of experts $n$ up to 4 or 5, beyond which the improvement stabilized.

% \begin{table}[t]
% \caption{Hyperparametric analysis in MoDE method with different number of DDPM.}\label{tab100}
% \centering
% \vskip 0.1mm \setlength{\tabcolsep}{1.5mm}
% \begin{tabular}{@{\extracolsep\fill}ccccccc}
% \toprule
% Dataset & Method & Top1 & Top5 &Acc  & AUC & EER\\
% \midrule
% \multirow{5}{*}{\makecell[c]{MS1M \\ (Masked)}} & MoDE(1)  & 89.2 & 95.0  & 99.2 & 0.9941  & 0.027 \\
% ~ & MoDE(2) & 89.2 & 94.8  & 99.2 & 0.9940  & 0.028  \\ 
% ~ & MoDE(3) & 89.0  & 95.0  & 99.2 & 0.9941  & 0.025 \\
% ~ & MoDE(4) & 88.0  & 95.0  & 99.2 & 0.9942  & 0.029 \\
% ~ & MoDE(5)  & 89.0  & 95.0  & 99.2 & 0.9940  & 0.027 \\
% \midrule
% \multirow{5}{*}{\makecell[c]{CelebA \\ (Masked)}} & MoDE(1)  & 29.9  & 44.5  & 80.5 & 0.9190  & 0.150  \\
% ~ & MoDE(2) & 31.2  & 46.0  & 80.6 & 0.9232  & 0.145  \\ 
% ~ & MoDE(3) & 31.5  & 46.2  & 81.3 & 0.9261  & 0.140  \\
% ~ & MoDE(4) & 32.1  & 46.1  & 81.7 & 0.9255  & 0.145  \\
% ~ & MoDE(5)  & 32.0  & 46.5  & 81.8 & 0.9254  & 0.146  \\
% \bottomrule
% \end{tabular}
% \end{table}

\begin{figure}[t]
\centering
\includegraphics[width=0.47\textwidth]{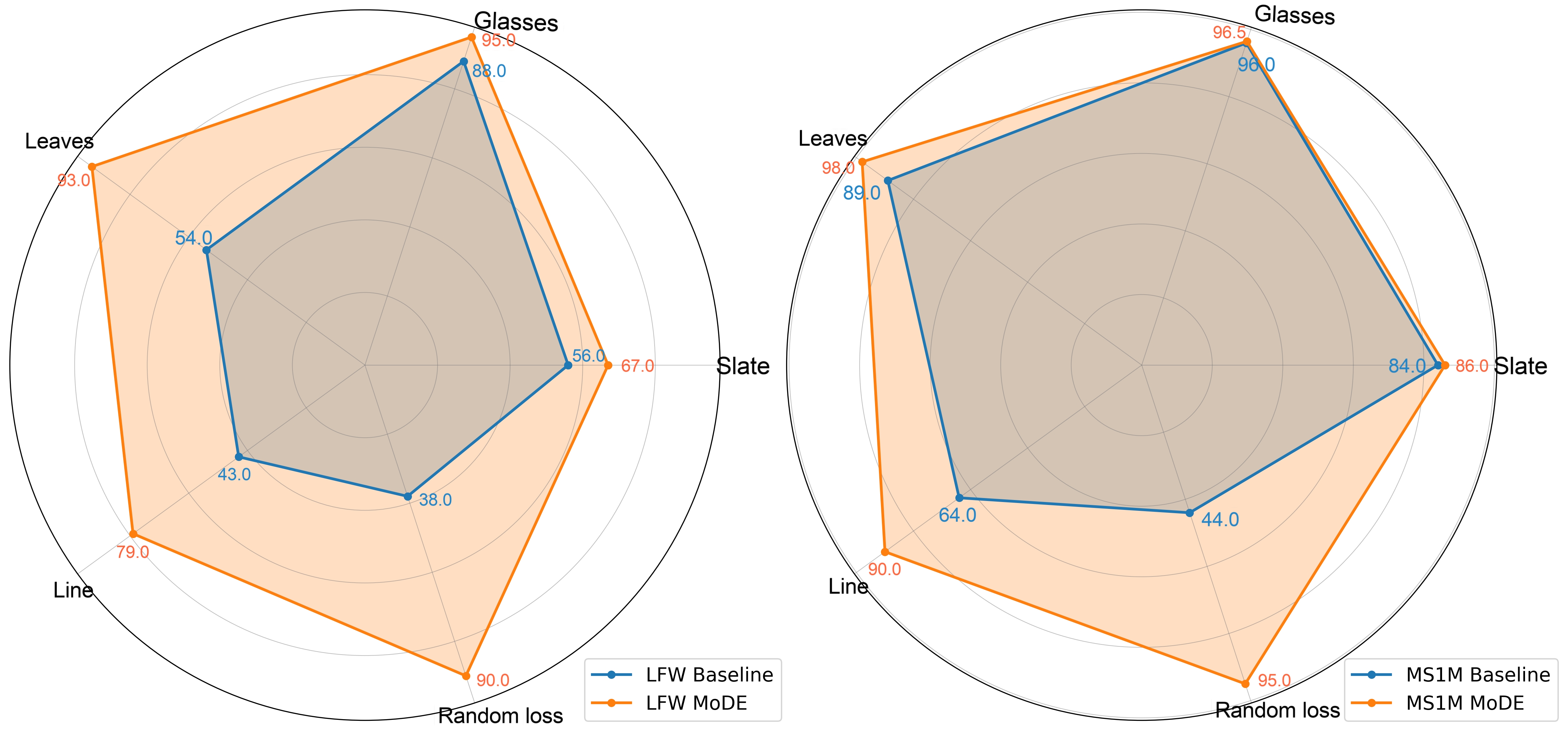}
% \vspace{-8pt}
\caption{An illustration of accuracy between different occlusions in LFW and MS1M. The results demonstrate the robustness of the MoDE approach across various occlusions.}
\label{fig8}
\end{figure}

% \begin{table}[t]
% \caption{Experimental results under different occlusions.}\label{tab77}
% \vspace{-1pt}
% \centering
% \vskip 0.1mm \setlength{\tabcolsep}{1mm}
% \resizebox{.95\columnwidth}{!}{
% \begin{tabular}{@{\extracolsep\fill}ccccccc}
% \toprule
% Dataset & Method & Slate & Glasses & Leaves & Line & Random loss\\ 
% \midrule
% \multirow{2}{*}{LFW} & baseline & 56.0 & 88.0 & 54.0 & 43.0 & 38.0\\
% ~ & MoDE & \textcolor{red}{67.0} & \textcolor{red}{95.0} & \textcolor{red}{93.0} & \textcolor{red}{79.0} & \textcolor{red}{90.0}\\
% \midrule
% \multirow{2}{*}{MS1M} & baseline & 84.0 & 96.0 & 89.0 & 64.0 & 44.0\\
% ~ & MoDE & \textcolor{red}{86.0} & \textcolor{red}{96.5} & \textcolor{red}{98.0} & \textcolor{red}{90.0} & \textcolor{red}{95.0}\\
% \toprule
% \end{tabular}
% }
% % \vspace{-3px}
% \end{table}

\noindent\textbf{Robustness Validation:} To assess the efficacy of MoDE on diverse occlusion types, we conducted experiments on the occluded MS1M and LFW datasets, using ArcFace as the baseline. Our results, as presented in Fig. \ref{fig8}, clearly show that MoDE outperforms the baseline method in recognizing occluded faces. 
Moreover, MoDE demonstrated superior robustness on various types of occlusion and excelled in handling fragmented or randomly dispersed occlusions such as Leaves, Random loss, and Lines.

\section{Conclusion}

In this paper, we propose an innovative method for occluded face recognition, the Mixtures of Diffusion Experts (MoDE), which employs multiple diffusion experts for reconstruction and utilizes ID-Gate for dynamic integration of expert predictions in the decision space. MoDE is a \textit{plug-and-play} module and exhibits stable and substantial improvements on most face recognition models. Excellent performance on both public and wild datasets with various occlusions demonstrates the robustness of MoDE. Furthermore, our approach presents a novel and promising framework for occluded face recognition, potentially inspiring further research in this domain.

\balance

\bibliographystyle{IEEEtran}

\end{document}